\documentclass[conference]{IEEEtran}
\ifCLASSINFOpdf
\else
\fi
\usepackage[tight,footnotesize]{subfigure}

\usepackage{graphicx}

\usepackage{caption}

\usepackage{url}

\usepackage{hyperref}

\DeclareRobustCommand{\IEEEauthorrefmark}[1]{\smash{\textsuperscript{\footnotesize #1}}}

\begin{document}

\title{Towards Space-to-Ground Data Availability for Agriculture Monitoring}

\author{\IEEEauthorblockN{George Choumos\IEEEauthorrefmark{1}$^*$, Alkiviadis Koukos\IEEEauthorrefmark{1}$^*$, Vasileios Sitokonstantinou\IEEEauthorrefmark{1,2} and Charalampos Kontoes\IEEEauthorrefmark{1}}

\IEEEauthorblockA{\IEEEauthorrefmark{1}National Observatory of Athens, IAASARS, BEYOND Center, Athens, Greece}

\IEEEauthorblockA{\IEEEauthorrefmark{2}Laboratory of Remote Sensing, National Technical University of Athens,  Athens, Greece\\
Email: \{g.choumos, akoukos, vsito, kontoes\}@noa.gr
}}


%


\setlength{\lineskip}{0pt}

\maketitle
\def\thefootnote{*}\footnotetext{These authors contributed equally to this work}

\begin{abstract}

The recent advances in machine learning and the availability of free and open big Earth data (e.g., Sentinel missions), which cover large areas with high spatial and temporal resolution, have enabled many agriculture monitoring applications. One example is the control of subsidy allocations of the Common Agricultural Policy (CAP). Advanced remote sensing systems have been developed towards the large-scale evidence-based monitoring of the CAP. Nevertheless, the spatial resolution of satellite images is not always adequate to make accurate decisions for all fields. In this work, we introduce the notion of space-to-ground data availability, i.e., from the satellite to the field, in an attempt to make the best out of the complementary characteristics of the different sources. We present a space-to-ground dataset that contains Sentinel-1 radar and Sentinel-2 optical image time-series, as well as street-level images from the crowdsourcing platform Mapillary, for grassland fields in the area of Utrecht for 2017. The multifaceted utility of our dataset is showcased through the downstream task of grassland classification. We train machine and deep learning algorithms on these different data domains and highlight the potential of fusion techniques towards increasing the reliability of decisions.

\begin{IEEEkeywords}
agriculture, crop type classification, satellite image time series, geo-tagged street-level pictures, data fusion
\end{IEEEkeywords}

\end{abstract}



%
\IEEEpeerreviewmaketitle

\section{Introduction}
The Common Agricultural Policy (CAP) of the European Union (EU) is changing fast. It aims at taking advantage of the recent advances in information technologies, machine learning and earth observation to shift towards the evidence-based monitoring of farmers' compliance with the rules. CAP Paying Agencies (PA) of each Member State have to move away from the currently employed confined and sample-based checks and adopt practices that will enable them to take informed decisions for the subsidy payments of all parcels. For that, the Area Monitoring System (AMS) is introduced. AMS refers to the systematic monitoring and assessment of agricultural activities and practices using Satellite data (e.g. Copernicus), joined with data from the Land Parcel Identification Systems (LPIS), but also other ancillary sources. 
The Copernicus Sentinel satellite missions provide optical and Synthetic Aperture Radar (SAR) data of high spatial and temporal resolution and have been extensively used for the agricultural monitoring and the CAP purposes \cite{lopez2021sentinel, sonobe2017assessing}. 

On the other hand, the main enabler towards the feasibility of this exhaustive monitoring, is widely considered to be Artificial Intelligence (AI). In particular, Machine Learning (ML) and Deep Learning (DL) pipelines are continuously being developed and integrated into the operational framework of the PAs \cite{lopez2021sentinel, schulz2021large, navarro2021operational, rousi2020semantically, sarvia2021possible}.
However, for wall-to-wall and perfectly accurate assessment on the compliance of farmers, as required by the new CAP guidelines, the data modality and spatial resolution (10 m - 60 m) of the Sentinel imagery alone is not enough. In fact, it is required to incorporate other sources to complement their weaknesses. 

Remote sensing systems usually produce a large volume of data due to the high spatial, spectral, radiometric and temporal resolutions needed for applications in agriculture monitoring \cite{sishodia2020applications}. 
Looking at the current, rapidly advancing, state-of-the-art in ML for Earth Observation (EO), data availability is growingly perceived to be of multi-dimensional nature, through the notion of data variability and data complementarity, on top of data volume. Thus, more and more heterogeneous data sources are collected and data fusion techniques are applied in an attempt to generate enhanced feature spaces from non-overlapping and complementary feature domains. In EO applications during the recent years, fusion of data in different altitude levels, such as UAV and high resolution satellite images, is very common \cite{zhao2019finer, zhou2021uav}.
However, to convert the available data into training datasets for ML/DL pipelines, we also require the relevant annotations, i.e., ground-truth labels. 
While satellite imagery is not in short supply, ground-level observations are hard to find and they lack consistency in terms of spatial and temporal availability. Apart from being difficult to acquire, such labels are also the most expensive data component to generate; not just in terms of monetary cost, but also in terms of time, workforce and the availability of expert knowledge for the annotation.

Going back to the new CAP monitoring requirements, PAs are facing challenges in their attempt to monitor compliance. Through the implementation of the AMS, the decisions that will be made for the subsidy allocations of all parcels have to be supported by evidence, either as a result of the predictions of ML/DL pipelines, or through photo-interpretation, which is a common process that the PAs undertake (in retrospect) for validation purposes and for cases of dispute resolution. Thus, the need for such a holistic approach regarding annotated data availability is becoming evident.

Driven by the above, we present an analysis-ready dataset, annotated with crop type labels, that includes both the space (Sentinel data) and ground (street-level images) domains. We formulate this framework of space and ground data availability using only open-access (Copernicus data and LPIS) and crowdsourced (Mapillary) data. We present a methodology and share the code for i) collecting, ii) processing to analysis-ready and iii) annotating (with crop labels) street-level images. Each street-level image is matched with a time-series of Sentinel-1 (S1) and Sentinel-2 (S2) data. We use our dataset to independently train traditional ML and state-of-the-art DL models on its space and ground data layers, and we discuss the capabilities of synergistic use by applying late fusion. All in all, we introduce the idea of space-to-ground data availability and offer an example curated dataset that the community is encouraged to work on using DL fusion models, on both satellite and street-level images, towards fully exploiting the complementarity of these data modalities for agriculture monitoring downstream tasks.

\section{Data availability for Remote Sensing applications}
Our initial approach to this multi-level (i.e., altitude, mode, angle of sensing etc.) data availability framework includes the two edges, i.e., space-level (low earth orbit) and ground-level. In each of these domains, the data outputs differ substantially in terms of a set of characteristics, such as the spatial and spectral resolution, the temporal availability, the viewing angle, the susceptibility to the weather conditions, and the nature of the conclusions that can be drawn through their use.

\subsection{Sources going from space to ground}
Space-level includes acquisitions from EO satellites starting from higher ranges of low earth orbits (e.g. Sentinel 1/2), and moving down to lower altitudes (e.g. SPOT6/7 and Planet's Doves). On the other edge, the ground-level comprises imagery and acquisitions coming from the street (e.g. Google Street View, Mapillary and Kakao), as well as in-situ geotagged photos from mobile phones \cite{kenny2021empathising}, which is a source that is already utilized for agricultural monitoring \cite{wang2020mapping} as well as for validation processes and evidence collection campaigns of the PAs. Between these two edges of space and ground data acquisitions, there is the potential to integrate other data sources from various domains and acquisition altitudes, with the most common being aerial photos \cite{zhang2013fusion} and images from Unmanned Aerial Vehicles (UAVs) \cite{maimaitijiang2020crop}.

\subsection{Data availability in agriculture monitoring}
\subsubsection{Space-level datasets}
Data availability on the space-level is quite extensive, with a growing number of EO satellites providing a consistent stream of data acquisitions over the globe. The most commonly used data sources on this level for the CAP monitoring purposes are the Sentinel and Landsat satellites, with their acquisitions becoming available as open-access data and on various processing levels. Thus, there is a plethora of available datasets with annotations for a wide range of downstream applications. BigEarthNet\cite{8900532} is a collection of 590,326 pairs of S1 and S2 image patches over Europe, annotated with multiple land cover classes. DENETHOR \cite{Kondmann2021DENETHORTD} is a publicly available analysis-ready benchmark dataset which combines Planet Fusion data, together with S1 radar and S2 optical data. It includes annotations of crop type labels, covering an area in Northern Germany. ZueriCrop\cite{turkoglu2021crop} is another case of a crop classification targeted dataset, containing S2 time series, along with ground truth labels for 116,000 fields over Zurich. Sen4AgriNet\cite{sen4agrinet} is a benchmark dataset based on S2 imagery, annotated with crop type labels, which contains 42,5 million parcels, thus, rendering it suitable for the training of DL architectures. Finally, CropHarvest \cite{tseng2021cropharvest} contains S1 and S2 time-series, meteorological and topographic data, together with crop type labels for 90,480 datapoints spanning all around the world.

\subsubsection{Ground-level datasets}
As previously described and, in contrast to the space-level datasets, securing availability on the ground-level is a challenging pursuit. This is both in terms of finding data relevant to the CAP monitoring domain, as well as in regards to finding or generating annotations. The main focus of street-level imagery in computer vision is commonly the identification of objects and features that are encountered in urban areas (e.g. cars, traffic signs etc.), driven by domains of application like self-driving automobiles. Thus, there is a general shortage in ground-level datasets from rural areas and, even more, a shortage in annotations of agricultural interest, with most of the available ones coming from time-consuming photo-interpretation efforts. Despite the challenges, some remarkable collections of ground-level datasets exist. iCrop\cite{Wu2021-jb} is a multiclass dataset of 34,117 road view images in China, annotated with crop type labels. CropDeep\cite{CropDeep2019} includes 31,147 in-situ images and 40,000 annotations related to plant species classification, captured in a variety of realistic conditions. However, to the authors' knowledge, the majority of the available annotated datasets on this level are of significantly smaller size, which renders the effective training of ML/DL models through their use more challenging. For example, Crop/Weed Field Image Dataset \cite{haug15}, comprises 60 images with vegetation masks and annotations for crops and weeds.

\subsubsection{Combining Space and Ground}
The combined use of datasets from different sources is a common approach in the literature for feature enhancement and augmentation of the training data volume. These combinations are applied in a variety of ways, like on the measurement level \cite{maimaitijiang2020crop}, feature level \cite{zhang2013fusion}, and decision level \cite{chen2020decision}. However, such datasets are usually limited to the space domain (e.g., S1 and S2 \cite{sitokonstantinou2021scalable}, Sentinel and Landsat-8 \cite{chen2020decision}, Sentinel and PlanetFusion \cite{Kondmann2021DENETHORTD} etc.).
Consequently, availability on the basis of fusing different domains (e.g. space and ground) is limited.

\subsection{Ground coverage with crowdsourced and open-access data}
Given the amount of ground-level data acquisitions that are required to match and fully cover a single space-level acquisition, we are driven towards the need to decentralize the effort of ground-level data collection and the exploitation of crowdsourcing platforms like Mapillary, Google Earth Pro and Kakao. However, while such platforms can offer significant volumes of ground-level data, they do come with certain particularities that need to be addressed before they can be considered analysis-ready. For instance, crowdsourced contributions may display discrepancies in the acquisition methodology. Also, there are no annotations available for most of the contributions, especially ones of agricultural interest. To counter this shortage of annotations, methodologies for their mass generation are capturing the interest of the research community. LPIS data, which contain georeferenced crop type labels coming from farmers' declarations, have already been used in DataCAP \cite{datacap} to map crop types to street-level images based on transformations to the coordinates of the image acquisition. Street2Sat\cite{street2sat} pipeline is aiming to transform geotagged street-level images to sets of georeferenced points that can be used as labels for satellite images. In \cite{dAndrimontSLI}, the authors developed a pipeline for the monitoring of crop phenology using DL architectures on street-level imagery, reference parcel data, and ground observations.

\section{The Dataset}

In this work, we introduce a multi-level, multi-sensor, multi-modal dataset annotated with grassland/not grassland labels for the monitoring of the CAP. We empower the transition towards the post-2020 CAP guidelines through the exploitation of the benefits offered by the dataset's space and ground components. Our methodology for data collection and annotation is presented, and the code is available at \url{https://github.com/Agri-Hub/Space2Ground}. We showcase the multi-faceted utility of our dataset, by applying state-of-the-art ML/DL architectures for grassland detection on data from the space and ground domains. On top of the single-domain results, we apply late-fusion (decision level) to underline the potential of harnessing our dataset's multi-modality.

\subsection{Data sources and data collection}
We constructed this dataset considering the two edges of data availability. On the space-level edge, we have included Sentinel-1 GRD and SLC products and Sentinel-2 multispectral imagery, acquired through the Copernicus Open Access Hub. Specifically, from Sentinel-1 we produced VV/VH backscatter and VV/VH coherence using the snappy library and from Sentinel-2, we acquired all bands except 1, 9 and 10. On the ground-level edge, we use the Mapillary crowdsourcing platform to include street-level images covering our area of interest. Mapillary data is available for anyone to explore under a CC-BY-SA license agreement. In order to interact with the latest version of the Mapillary API (v4), we developed a Python library. The scripts for downloading imagery from Mapillary, based on a specific time range for a specific area can be found on GitHub (\url{https://github.com/Agri-Hub/Callisto/tree/main/Mapillary}).
For the annotation of these datasets with agricultural crop type labels, we use the Dutch LPIS data which are openly available through the National Georegister website \cite{brp_data} under no limitations for access and use.

Grassland is the most dominant crop type in the Netherlands. In our AOI, there are 55,039 parcels in total and more than 80\% have been declared as Grassland. We excluded instances without clear spectral signatures or irrelevant to CAP purposes (e.g. forests, borders adjacent to arable land, etc.) and heavily impacted by cloud coverage, ending up with a total of 37,041 parcels, out of which 31,350 (84.6\%) are grasslands.

\subsection{Methodology of annotation}
To annotate our data we use the LPIS dataset. It consists of shapefiles with geo-referenced parcel geometries, that are assigned with crop type labels. As Sentinel data are also geo-referenced, their annotation with these labels is straightforward.
The generation of labels for the street-level images, though, is a more challenging task. Mapillary includes information about the coordinates of the sensor in each image capture and a compass angle to specify the direction towards which it is facing. To generate annotations for the street-level images, we fuse them with the LPIS data by applying transformations to the acquisition coordinates following the methodology developed in \cite{datacap}. Using this methodology we end up with up to 2 labels for each single image (one for the left and one for the right side of the field of view).

\begin{figure*}[!ht]
\centering
\includegraphics[width = \textwidth]{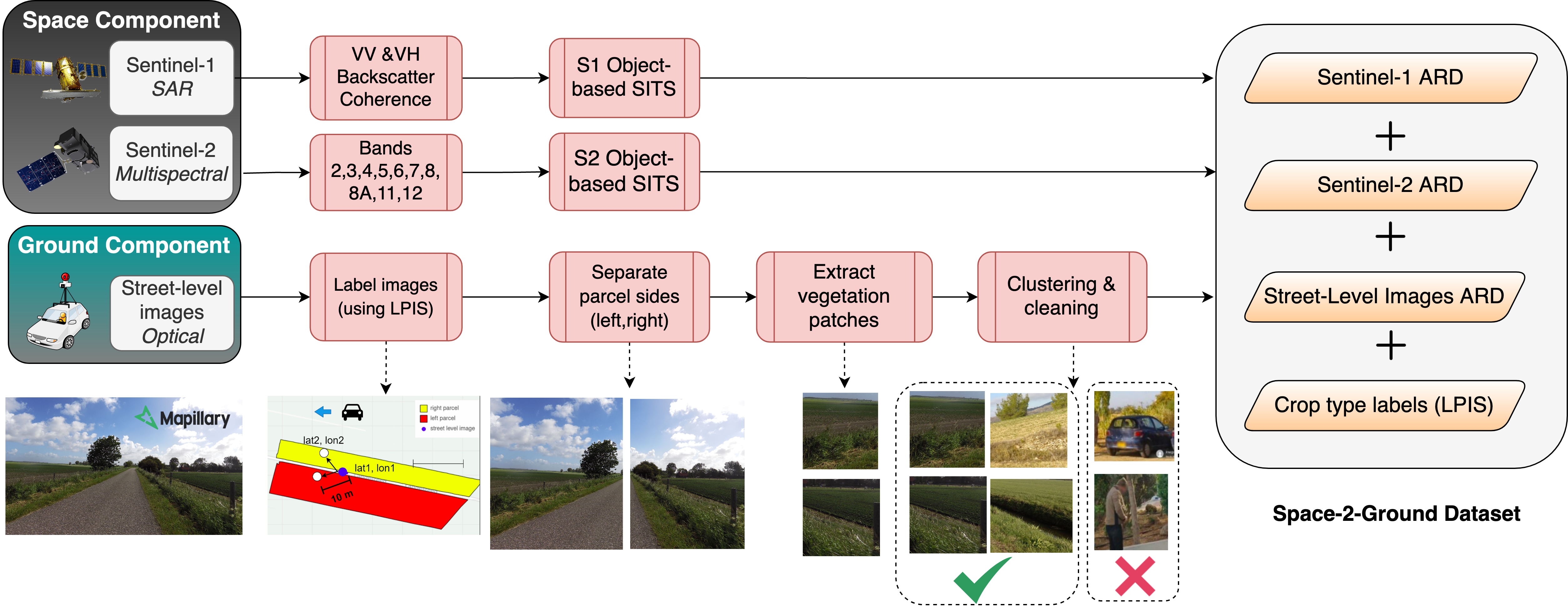}
\caption{Methodology for the dataset creation. The Space component consists of S1 and S2 object-based time series data. The Ground component consists of street-level imagery acquired through the Mapillary platform. Transformations to the acquisition coordinates are applied to attach LPIS crop type labels to left and right parcels of each image. Each parcel side is isolated and vegetation patches are extracted. A final cleaning step is applied through k-means clustering and photo-interpretation.}
\label{fig:methodology}
\end{figure*}

On top of this approach, we apply additional steps to further refine the images and improve the quality of the annotations. Initially, we split each image in 2 halves, separating the labels of each side accordingly (with a front-facing acquisition, each side corresponds to a different parcel). Then, to isolate and extract vegetation, we keep the 30\% top leftmost or rightmost part (depending on the side) with regards to width, and the 20\% to 50\% part with regards to height. Through these steps, we end up with 36,985 image patches, which are resized to (260, 260). Subsequently, we filter out ones that are irrelevant or contain noise (eg. cars, infrastructure, roads, etc.) in an unsupervised way. In particular, we use a VGG16 pretrained network to extract a representation of each image. Then, we perform Principal Component Analysis on these representations and we keep the 100 most important features. Finally, we fit a k-means model to the aforementioned data to group them into 200 clusters, and we keep the 50 most useful ones through visual interpretation. After this filtering step, we end up with a refined set of 10,102 annotated image patches.

\subsection{Dataset structure}

On the space-level, we calculate mean values of Sentinel-1 and Sentinel-2 data for each parcel. Sentinel-1 data are also aggregated on a monthly basis. Consecutively, we create training and test sets (80\%-20\%), and store them in csv format. The first two columns correspond to the the crop label (Grassland/Non-Grassland) and the unique identifier of the parcel, while the rest of the columns represent the mean values of the Sentinel features.
On the ground-level, the street level images are also split into training and test data (80\%-20\%), and they are stored in different directories based on their class label. The image names are in form of $\{imageID\}\_\{direction\}$, where $\{imageID\}$ corresponds to the unique identifier of the image and $\{direction\}$ to the direction in which it has been cropped.

\subsection{Dataset utility \& downstream tasks}
Through the provision of an object-level mapping between its different data sources, our dataset facilitates the creation of pipelines that harness it in a multitude of ways. Each single data source can be used separately for the training of ML/DL pipelines. Various modes of data fusion are supported, like measurement fusion of same-level data, feature fusion, and late fusion of models trained separately on different data layers and levels. The object-level mapping also allows for any available annotations to be broadcast to all data sources. 

The ground component comes with a set of additional use cases. Ground data can work as material for validation of the ML/DL predictions and for the creation of ground-truth annotations. They are also suitable for other photo-interpretation tasks. For instance, PAs commonly undertake photo interpretation effort, for desk inspections, known as desk on-the-spot-checks (OTSC), and for resolution of disputes. Models trained on ground-level data can be mounted on ground sensors (cars, mobile phones, UAVs, etc) to support inference at the edge. Last but not least, Space-to-Ground data availability is greatly facilitating the creation of synthetic data through the use of generative architectures, like Generative Adversarial Networks (GAN), which are already being applied, particularly for the translation of data between the domains of satellite and street-level imagery \cite{sat2street,street2sat}.

\section{Benchmarking}
\label{sec:results}

We evaluated the performance of a Random Forest (RF), a Support Vector Machine (SVM) model and three DL models on our dataset for the task of grassland classification using S1 and S2 data as input. RF and SVM are commonly used for remote sensing tasks, thus we included it
as a benchmark to compare with the DL models. The different DL architectures consist of a temporal Convolutional Neural Network (CNN) \cite{Pelletier2019Temporal}, an LSTM with 64 hidden units and the same LSTM model with an additional attention mechanism. Table \ref{tab:metrics} depicts the performance of these models on the test dataset. 

\begin{table}[!ht]
\caption{Performance metrics for grassland classification using different models.}
\label{tab:metrics}
\centering
\begin{tabular}{|c|c|c|c|c|c|}
\hline
\textbf{Method}   & \textbf{SVM} & \textbf{RF}  & \textbf{TempCNN}  & \textbf{LSTM} & \textbf{LSTM+Attention} \\ \hline
\textbf{Accuracy} & 93.69\%  & 94.68\%  & 95.22\%  & 95.14\%  & 95.20\%  \\ \hline
\textbf{F1 score} & 85.22\%  & 88.08\%  & 89.96\%  & 89.85\%  & 90.05\% \\ \hline
\end{tabular}
\end{table}

Apart from the space-level data, we also evaluated the performance of advanced CNNs (e.g. ResNet, EfficientNet, VGG, Inception) on the street level images. We experimented with pretrained (on ImageNet) models and the best performance was achieved by the InceptionV3, with an overall accuracy of 85\%. 
Therefore, the application of fusion at the decision level is enabled, by exploiting the predictions derived from both the space and ground data components. In our case, the combination of the outputs of these different models, results to marginal improvement of the overall accuracy. However, we notice significant enhancement in terms of the confidence of the final prediction and thus the reliability of the final decision.

\section{Conclusions and Future work}

In this work we presented the first dataset that includes Sentinel-1, Sentinel-2, and street-level images, matched through the use of geo-referenced crop type labels (LPIS). We focused on grasslands for a Dutch AOI around Utrecht and we used openly accessible and crowdsourced data sources.
The code implementation of our methodology is shared, which renders it reproducible, transferable and extensible for the community to utilize and build on top.
Performance of the grassland classification models is impacted by limitations related to the quantity and quality of the street-level images. 
With regards to quantity, larger areas of interest and wider time-frames must be considered to increase the number of available images. To this direction, PAs can also exploit the current operational framework of their daily inspections, and mount cameras on the field inspectors' vehicles to automatically capture imagery during their field visits.
From the image quality perspective, pointing cameras on the side can dramatically increase the percentage of vegetation per image and cover larger portions of the parcels. In addition, vegetation extraction can be improved through the application of semantic segmentation and creation of appropriate masks \cite{seamless_scene_segmentation}.
The potential for application of various fusion approaches will be further explored, in order to highlight the interoperability and complementarity of our dataset's layers and domains.


\section*{Acknowledgment}

This work has been supported by the CALLISTO and e-shape projects, funded by EU's Horizon 2020 research and innovation programme under grant agreements No. 101004152 and No. 820852.



\bibliographystyle{IEEEtran}
\bibliography{refs}
%
%
%

\end{document}